\documentclass[conference]{IEEEtran}
\IEEEoverridecommandlockouts
\usepackage{cite}
\usepackage{amsmath,amssymb,amsfonts}
\usepackage{algorithm}      
\usepackage{algpseudocode}  

\algrenewcommand\algorithmicrequire{\textbf{Input:}}
\algrenewcommand\algorithmicensure{\textbf{Output:}}
\usepackage{booktabs}   
\usepackage{multirow}   
\usepackage{pifont}     

\usepackage{graphicx}
\usepackage{textcomp}
\usepackage{xcolor}

\usepackage{tikz}
\usetikzlibrary{arrows.meta, positioning, calc}
\usepackage{float}

\def\BibTeX{{\rm B\kern-.05em{\sc i\kern-.025em b}\kern-.08em
    T\kern-.1667em\lower.7ex\hbox{E}\kern-.125emX}}
\begin{document}

\title{Scalable Learning in Structured Recurrent Spiking Neural Networks without Backpropagation\\
}

\author{\IEEEauthorblockN{1\textsuperscript{st} Bo Tang}
\IEEEauthorblockA{\textit{Department of Electrical and Computer Engineering} \\
\textit{Worcester Polytechnic Institute}\\
Worcester, USA \\
btang1@wpi.edu}
\and
\IEEEauthorblockN{2\textsuperscript{nd} Weiwei Xie}
\IEEEauthorblockA{\textit{Department of Electrical and Computer Engineering} \\
\textit{Worcester Polytechnic Institute}\\
Worcester, USA \\
wxie4@wpi.edu}

}

\maketitle

\begin{abstract}
Spiking Neural Networks (SNNs) provide a promising framework for energy-efficient and biologically grounded computation; however, scalable learning in deep recurrent architectures with sparse connectivity remains a major challenge. In this work, we propose a structured multi-layer recurrent SNN architecture composed of locally dense recurrent layers augmented with sparse small-world long-range projections to a readout population. The long-range connectivity is largely fixed, preserving routing efficiency and hardware scalability, while synaptic adaptation is performed using strictly local plasticity mechanisms. To enable supervised learning without backpropagation or surrogate gradients, we introduce a biologically motivated learning framework that combines: (i) population-based winner-take-all (WTA) teaching signals at the output layer, (ii) fixed random broadcast alignment feedback pathways, and (iii) low-dimensional modulatory neuron populations that gate synaptic updates through three-factor learning rules with eligibility traces. This design supports deep recurrent computation with sparse global communication and purely local synaptic updates. We analyze the algorithmic properties, computational complexity, and hardware feasibility of the proposed approach, and demonstrate stable learning and competitive performance on benchmark classification tasks. The results highlight the potential of structured recurrence and neuromodulatory learning to enable scalable, hardware-compatible SNN training beyond gradient-based methods.
\end{abstract}

\begin{IEEEkeywords}
Spiking neural networks, local plasticity, locally dense recurrent, sparse small-world long-range projection, winner-take-all teaching
\end{IEEEkeywords}

\section{Introduction}

Spiking Neural Networks (SNNs) constitute a biologically inspired computational paradigm in which information is represented and processed through sparse, asynchronous spike events \cite{Maass1997}. This event-driven representation enables natural compatibility with neuromorphic hardware platforms, where computation and communication are tightly coupled and energy consumption is dominated by spike activity rather than dense numerical operations \cite{Indiveri2015,Boahen2017}. As a result, SNNs offer significant advantages in energy efficiency, latency, and scalability over conventional artificial neural networks when deployed on dedicated hardware. Despite these advantages, learning in large-scale and recurrent SNNs remains a fundamental challenge.

A central difficulty arises from the non-differentiable nature of spike generation, which complicates the application of gradient-based optimization methods. Surrogate gradient techniques have been proposed to enable gradient descent by approximating derivatives through spike events \cite{Neftci2019,Bellec2020}. While effective in many settings, these approaches typically rely on global error propagation, dense feedback pathways, and precise temporal synchronization, all of which are difficult to reconcile with biological constraints and hardware limitations. In recurrent networks, these issues are further compounded by long temporal credit assignment paths, leading to increased training instability and computational cost.

To mitigate these challenges, many existing SNN architectures adopt either strictly feedforward structures \cite{Diehl2015} or untrained random recurrent reservoirs with trainable readout layers \cite{Maass2002}. Although such designs simplify learning and improve stability, they significantly diverge from the organization of biological neural circuits. Cortical networks are characterized by dense local recurrence, sparse long-range projections, small-world connectivity patterns \cite{Watts1998}, and modulatory control of synaptic plasticity \cite{roelfsema2018control}. Ignoring these structural and functional properties limits both the expressiveness of SNN models and their relevance to neuromorphic systems.

Recent biologically motivated learning frameworks have explored alternatives to backpropagation, including local three-factor learning rules with eligibility traces \cite{Gerstner2018}, broadcast alignment \cite{Lillicrap2016}, reward-modulated plasticity \cite{roelfsema2018control}, and population-based teaching signals \cite{Bellec2020}. However, most prior work has focused on shallow architectures, dense connectivity, or purely feedforward networks, leaving open the question of how deep recurrent SNNs can be trained in a stable and scalable manner using sparse communication and local learning alone. In particular, the interaction between structured recurrence, sparse global routing, and modulatory learning signals remains insufficiently explored.

In this work, we address these gaps by proposing a structured multi-layer recurrent SNN architecture and an associated learning framework guided by the following principles:
\begin{itemize}
\item \textbf{Locally dense recurrent computation}: Each layer forms a structured recurrent microcircuit that supports rich temporal dynamics and memory through local interactions.
\item \textbf{Sparse small-world long-range communication}: A limited number of long-range projections connect recurrent layers directly to the readout population, enabling efficient information routing while preserving sparsity.
\item \textbf{Low-bandwidth feedback}: Error information is transmitted through fixed random broadcast pathways rather than layer-wise backpropagation, reducing feedback bandwidth and routing complexity.
\item \textbf{Modulatory control of plasticity}: Synaptic updates are gated by low-dimensional modulatory neuron populations, consistent with neuromodulatory mechanisms observed in biological neural systems.
\item \textbf{Hardware-friendly learning}: Learning relies exclusively on local eligibility traces and scalar modulatory signals, avoiding surrogate gradients and global synchronization.
\end{itemize}

We demonstrate that this combination enables deep recurrent SNNs with thousands of neurons to be trained on supervised classification tasks using purely local learning rules, achieving stable performance while remaining compatible with the constraints of neuromorphic and FPGA-based hardware platforms.

\section{Background}

\subsection{Spiking Neuron Models}

SNNs differ from conventional artificial neural networks in that information is transmitted through discrete spike events rather than continuous-valued activations. Among the various neuron models proposed in the literature, the leaky integrate-and-fire (LIF) neuron represents a widely used compromise between biological realism and computational efficiency \cite{GerstnerKistler2002}. In its discrete-time form, the membrane potential \( V_i(t) \) of neuron \( i \) evolves according to
\begin{equation}
V_i(t+1) = \alpha V_i(t) + I_i(t) - V_{\mathrm{th}} S_i(t),
\end{equation}
where \( \alpha \in (0,1) \) is a leak factor, \( I_i(t) \) denotes synaptic input current, \( V_{\mathrm{th}} \) is a firing threshold, and \( S_i(t) \in \{0,1\} \) is the emitted spike determined by a thresholding operation. This model captures key temporal properties of biological neurons while remaining suitable for large-scale simulation and hardware implementation.

\subsection{Recurrent Spiking Networks and Temporal Credit Assignment}

Recurrent connectivity enables SNNs to represent temporal context and internal memory, but also introduces challenges related to stability and credit assignment over time. Classical recurrent SNN models include randomly connected reservoirs, such as liquid state machines \cite{Maass2002} and echo-state networks adapted to spiking dynamics \cite{Jaeger2001}. In these approaches, recurrent weights are fixed and only readout connections are trained, avoiding instability but limiting representational adaptability.

Training recurrent SNNs with learnable synapses is substantially more difficult. Gradient-based approaches typically rely on backpropagation through time combined with surrogate gradients to approximate derivatives of spike events \cite{Neftci2019,Bellec2020}. While effective, these methods require dense feedback pathways and global synchronization, posing challenges for biological plausibility and neuromorphic hardware deployment.

\subsection{Local Synaptic Plasticity and Eligibility Traces}

Biologically grounded alternatives to backpropagation are commonly formulated in terms of local synaptic plasticity rules. Classical spike-timing-dependent plasticity (STDP) updates synapses based on precise pre- and postsynaptic spike timing, but on its own is insufficient to solve supervised learning tasks \cite{BiPoo1998}. More powerful learning mechanisms are captured by three-factor learning rules, in which synaptic changes depend on presynaptic activity, postsynaptic activity, and an additional modulatory signal conveying task-related information \cite{Fremaux2016}.

Eligibility traces provide a mechanism for bridging temporal delays between neural activity and delayed feedback signals \cite{Gerstner2018}. Each synapse maintains a local eligibility variable that accumulates correlations between pre- and postsynaptic spikes and decays over time. When combined with a modulatory signal, eligibility traces enable temporally extended credit assignment using strictly local information, making them well suited for recurrent SNNs and event-driven hardware.

\subsection{Broadcast Alignment and Low-Bandwidth Feedback}

Broadcast alignment has emerged as an alternative to precise error backpropagation, demonstrating that fixed random feedback weights can support effective learning in deep networks \cite{Lillicrap2016}. In the context of SNNs, broadcast alignment allows error-related signals to be distributed to hidden layers without requiring symmetric forward and backward weights or dense feedback connectivity. This approach significantly reduces feedback bandwidth requirements and aligns well with biological and hardware constraints.

\subsection{Winner-Take-All Teaching Signals}

Winner-take-all (WTA) mechanisms are ubiquitous in biological neural circuits and have been widely used in both rate-based and spiking models for decision making and competition \cite{Maass2000}. In supervised settings, population-based WTA teaching signals provide a sparse and sign-consistent error representation that depends only on the relative ordering of output activations rather than their precise magnitudes. Such signals avoid the need for probabilistic normalization and integrate naturally with three-factor plasticity and modulatory learning frameworks.

Together, these components, including recurrent spiking dynamics, local plasticity with eligibility traces, broadcast alignment feedback, and WTA-based teaching, form the conceptual foundation for the structured recurrent learning framework proposed in this work.

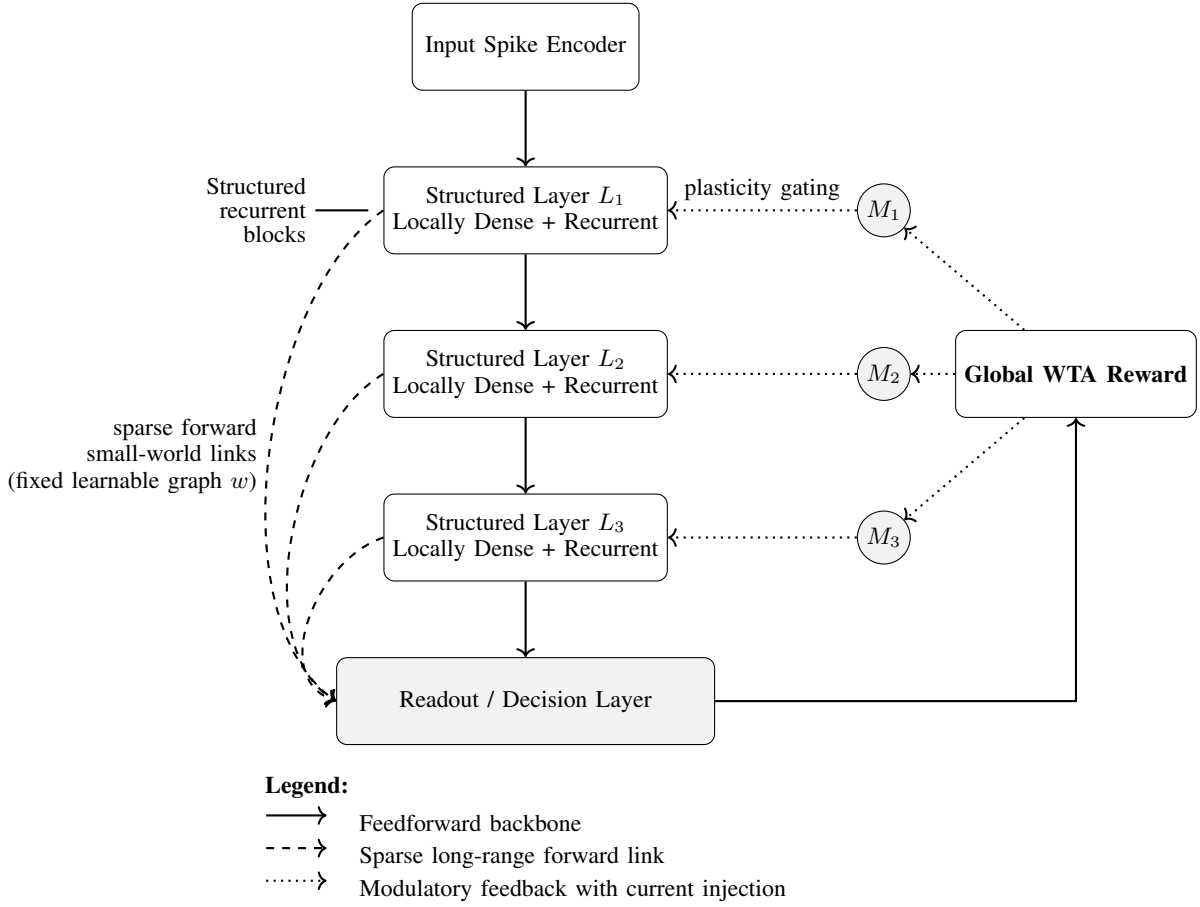
\begin{figure*}[hbt]
\centering
\begin{tikzpicture}[
    font=\small,
    layer/.style = {draw, rounded corners, minimum width=3.0cm, minimum height=1.15cm, align=center},
    readout/.style = {draw, rounded corners, minimum width=5.0cm, minimum height=1.15cm, align=center, fill=gray!10},
    modpop/.style = {draw, circle, minimum size=7mm, inner sep=0pt, fill=gray!10},
    arrow/.style = {->, thick},
    sparse/.style = {->, thick, dashed},
    recur/.style = {->, thick, loop right},
    modsig/.style = {->, thick, dotted},
    inj/.style = {->, thick, dotted},
    node distance=1cm
]

\node[layer] (input) {Input Spike Encoder};

\node[layer, below=of input] (l1)
{Structured Layer $L_1$\\Locally Dense + Recurrent};

\node[layer, below=of l1] (l2)
{Structured Layer $L_2$\\Locally Dense + Recurrent};

\node[layer, below=of l2] (l3)
{Structured Layer $L_3$\\Locally Dense + Recurrent};

\node[readout, below=of l3] (out)
{Readout / Decision Layer};

\draw[arrow] (input) -- (l1);
\draw[arrow] (l1) -- (l2);
\draw[arrow] (l2) -- (l3);
\draw[arrow] (l3) -- (out);

\draw[recur] (l1);
\draw[recur] (l2);
\draw[recur] (l3);

\draw[sparse] (l1.west) .. controls +(-1.6,-1.1) and +(-1.6,1.1) ..
node[midway, right, xshift = -102, align=right]
{\small sparse forward\\\small small-world links\\\small (fixed learnable graph $w$)} (out.west);

\draw[sparse] (l2.west) .. controls +(-1.2,-0.7) and +(-1.2,0.7) .. (out.west);
\draw[sparse] (l3.west) .. controls +(-0.9,-0.3) and +(-0.9,0.3) .. (out.west);

\node[layer, align=center, right=3.8cm of l2] (R)
{\textbf{Global WTA Reward}};

\node[modpop, right=2.5cm of l1] (m1) {$M_1$};
\node[modpop, right=2.5cm of l3] (m3) {$M_3$};
\node[modpop, right=2.5cm of l2] (m2) {$M_2$};

\draw[modsig] (R) -- (m1);
\draw[modsig] (R) -- (m3);
\draw[modsig] (R) -- (m2);

\draw[modsig] (m1.west) -- node[midway, above]
{\small plasticity gating} (l1.east);
\draw[modsig] (m3.west) -- (l3.east);
\draw[modsig] (m2.west) -- (l2.east);

\coordinate (midleft) at ($(l1.west)+(-0.9,0)$);
\draw[thick] (midleft) -- ($(l1.west)+(-0.2,0)$);
\node[align=right, anchor=east] at (midleft)
{\small Structured\\[-1pt]\small recurrent\\[-1pt]\small blocks};

\draw[->, thick] (out.east) -| (R.south);

\node[below=0.3cm of out, align=left] (legend) {
\textbf{Legend:}\\[3pt]
\begin{tabular}{@{}ll@{}}
\raisebox{0.25em}{\tikz{\draw[arrow] (0,0)--(0.8,0);}} & Feedforward backbone \\
\raisebox{0.25em}{\tikz{\draw[sparse] (0,0)--(0.8,0);}} & Sparse long-range forward link \\
\raisebox{0.25em}{\tikz{\draw[inj] (0,0)--(0.8,0);}} & Modulatory feedback with current injection \\
\end{tabular}
};

\end{tikzpicture}

\caption{\textbf{Our proposed stacked structured recurrent SNN with sparse small-world connectivity and modulatory feedback.} Each layer forms a locally dense recurrent microcircuit. Sparse long-range forward projections following a fixed small-world topology connect recurrent layers directly to the readout population and contribute to the output signal. Feedback signals are broadcast from the readout and integrated by low-dimensional modulatory neuron populations, which gate synaptic plasticity in selected layers through three-factor learning rules. No feedback currents are injected into neuronal membrane dynamics. Synaptic weights are trained on a fixed connectivity graph, with optional limited structural rewiring at later training stages.}
\label{fig:architecture-vertical}
\end{figure*}

\section{Proposed Architecture and Learning Framework}

We propose a structured multi-layer recurrent spiking neural network (SNN) that combines locally dense recurrent computation, sparse small-world long-range connectivity, and biologically motivated learning mechanisms. The overall architecture is illustrated in Fig.~\ref{fig:architecture-vertical}.

\subsection{Network Architecture}

The network consists of $K$ vertically stacked structured recurrent layers followed by a readout population. Each layer $L_k$ contains $N_k$ spiking neurons and forms a locally dense recurrent microcircuit. Feedforward projections connect consecutive layers, while each recurrent layer also maintains sparse long-range projections directly to the readout population. These long-range connections follow a small-world topology and enable efficient global information routing without dense connectivity.

Let $L_k$ denote the $k$-th layer. The total synaptic input current received by neuron $i$ in layer $k$ at time $t$ is given by
\begin{equation}
I_i^k(t) =
\sum_{j \in L_k} w_{ij}^{\mathrm{rec}} S_j^k(t)
+ \sum_{j \in L_{k-1}} w_{ij}^{\mathrm{ff}} S_j^{k-1}(t)
+ \sum_{j \in L_k} w_{ij}^{\mathrm{lr}} S_j^k(t),
\end{equation}
where $w^{\mathrm{rec}}$ denotes recurrent synapses within the layer, $w^{\mathrm{ff}}$ feedforward synapses from the preceding layer, and $w^{\mathrm{lr}}$ sparse long-range synapses projecting to the readout population.

Long-range connectivity graphs are mostly fixed during training to preserve routing efficiency and hardware scalability. Synaptic weights are adapted using local plasticity rules, while optional limited structural rewiring can be enabled at later training stages to prune weak connections and sample new ones under fixed fan-in and fan-out constraints.

\subsection{Neuron Dynamics}

Each neuron is modeled as a discrete-time leaky integrate-and-fire (LIF) unit. Let $V_i(t)$ denote the membrane potential of neuron $i$ at time step $t$, and let $S_i(t) \in \{0,1\}$ denote its spike output. The neuron dynamics are given by
\begin{align}
V_i(t+1) &= \alpha V_i(t) + I_i(t) - V_{\mathrm{th},i}(t)\, S_i(t), \\
S_i(t) &= H\!\left(V_i(t) - V_{\mathrm{th},i}(t)\right),
\end{align}
where $\alpha \in (0,1)$ is a leak factor controlling membrane decay, $V_{\mathrm{th},i}(t)$ is the firing threshold, $I_i(t)$ denotes the total synaptic input current, and $H(\cdot)$ is the Heaviside step function. When a neuron emits a spike, its membrane potential is reset through the subtraction of the threshold term.

To maintain stable firing activity across layers, we employ a slow homeostatic threshold adaptation mechanism. The firing threshold of each neuron evolves according to
\begin{equation}
V_{\mathrm{th},i}(t+1) = V_{\mathrm{th},i}(t) + \eta_{\theta} \bigl( \bar{S}_i(t) - S_{\mathrm{target}} \bigr),
\end{equation}
where $\bar{S}_i(t)$ is an exponentially smoothed estimate of the neuron's firing rate, $S_{\mathrm{target}}$ is a desired target firing rate, and $\eta_{\theta}$ controls the adaptation timescale. This mechanism prevents runaway excitation and promotes balanced activity in recurrent layers.

\subsection{Winner-Take-All Output Teaching Signal}

The readout population consists of $C$ output neurons corresponding to $C$ classes. Each recurrent layer projects sparsely to the readout population: only a fraction $\rho_{\mathrm{out}} \in (0,1]$ of all possible long-range connections is instantiated, defining the \emph{readout density}. The readout activation vector
\[
z = (z_1, \ldots, z_C) \in \mathbb{R}^C
\]
is computed from spike counts accumulated over the simulation window. Let $y \in \{1,\ldots,C\}$ denote the target class label.

Instead of optimizing a differentiable loss, we employ a population-based winner-take-all (WTA) teaching signal. The predicted class is
\begin{equation}
c^* = \arg\max_{c} z_c,
\end{equation}
and a margin violation is defined as
\begin{equation}
m = \max\bigl(0,\, \gamma + z_{c^*} - z_y \bigr),
\end{equation}
where $\gamma > 0$ is a fixed margin parameter. The resulting output error signal is
\begin{equation}
\delta_c =
\begin{cases}
+m, & c = c^* \neq y, \\
-m, & c = y \neq c^*, \\
0,  & \text{otherwise}.
\end{cases}
\end{equation}
This teaching signal depends only on the relative ordering of readout activations and does not require softmax normalization, probability estimates, or gradient computation.

\subsection{Broadcast Alignment and Modulatory Populations}

The output error vector $\boldsymbol{\delta} \in \mathbb{R}^C$ is broadcast to each recurrent layer using a fixed random feedback matrix
\[
B \in \mathbb{R}^{C \times N_k},
\]
where $N_k$ is the number of neurons in layer $L_k$. The broadcast feedback received by neuron $i \in L_k$ is given by
\begin{equation}
\tilde{\delta}_i(t) = \sum_{c=1}^{C} \delta_c(t)\, B_{ci}.
\end{equation}

Rather than acting directly on membrane potentials, this feedback is integrated by a low-dimensional modulatory population associated with each layer. Let $m_k(t) \in \mathbb{R}^M$ denote the modulatory state of layer $k$, where $M \ll N_k$. The modulatory dynamics are defined as
\begin{align}
u_k(t) &= \frac{1}{N_k} \sum_{i \in L_k} \tilde{\delta}_i(t), \\
m_k(t+1) &= \lambda_m m_k(t) + W_{e \rightarrow m} u_k(t), \\
L_i(t) &= \tanh\!\bigl( (W_{m \rightarrow h} m_k(t))_i \bigr),
\end{align}
where $\lambda_m \in (0,1)$ is a modulatory decay factor and $W_{e \rightarrow m}$, $W_{m \rightarrow h}$ are fixed random projection matrices. The resulting modulatory signal $L_i(t)$ gates synaptic plasticity but does not encode precise gradient information, reflecting low-bandwidth neuromodulatory signaling.

\subsection{Eligibility Traces and Synaptic Plasticity}

Synaptic plasticity is governed by local eligibility traces. Each synapse from presynaptic neuron $j$ to postsynaptic neuron $i$ maintains a trace $e_{ij}(t)$ that evolves as
\begin{equation}
e_{ij}(t+1) = \lambda_e e_{ij}(t) + S_j(t)\, S_i(t),
\end{equation}
where $\lambda_e \in (0,1)$ controls the decay timescale, and $S_j(t)$ and $S_i(t)$ denote pre- and postsynaptic spike events.

When modulatory feedback is present, synaptic weights are updated according to a three-factor learning rule:
\begin{equation}
\Delta w_{ij}(t) = -\eta\, e_{ij}(t)\, L_i(t),
\end{equation}
where $\eta$ is the learning rate. Only synapses in layers receiving modulatory feedback are plastic; all others remain fixed.

Readout synapses are trained separately using a local delta rule,
\begin{equation}
\Delta W^{\mathrm{out}} = -\eta_{\mathrm{out}}\, S^\top \boldsymbol{\delta},
\end{equation}
where $S$ denotes the vector of spike counts from the presynaptic population and $\eta_{\mathrm{out}}$ is the readout learning rate.

\subsection{Training Procedure and Discussion}

Algorithm~\ref{alg:training} summarizes the training procedure of the proposed structured recurrent SNN. Training proceeds in discrete epochs and mini-batches, where each input sample is first encoded into a sequence of spike trains and processed by the recurrent network for a fixed number of timesteps. During this forward simulation, neuron states evolve according to leaky integrate-and-fire dynamics, and local eligibility traces are updated online based solely on pre- and postsynaptic spike events.

After the forward pass, network outputs are computed from spike-count activations in the readout population. A winner-take-all (WTA) teaching signal is then generated by comparing the predicted class with the target label, producing a sparse, sign-consistent error vector. Importantly, this teaching signal depends only on the relative ordering of readout activations and does not require probability normalization or gradient computation.

The output error is used in two distinct ways. First, readout synapses are updated using a local delta rule that directly adjusts long-range projections to improve classification decisions. Second, the same error signal is broadcast to recurrent layers through fixed random feedback matrices. Rather than acting directly on neuron dynamics, this feedback is integrated by low-dimensional modulatory populations that produce scalar gating signals. These modulatory signals determine when and where synaptic plasticity is enabled, without conveying precise error magnitudes.

Synaptic updates within recurrent layers follow a three-factor learning rule combining eligibility traces, modulatory gating, and a fixed learning rate. This separation between neural computation and learning signals ensures that recurrent dynamics remain stable during inference, while plasticity is selectively enabled during learning. A slow homeostatic mechanism further regulates neuronal firing rates by adapting firing thresholds, preventing runaway excitation or quiescence.

Overall, the training algorithm avoids backpropagation, surrogate gradients, and backpropagation-through-time. All updates are local, event-driven, and compatible with sparse communication constraints. While this design does not optimize an explicit global loss function, it enables stable learning in deep recurrent architectures with fixed sparse connectivity, making it well suited for neuromorphic and hardware-constrained implementations.

\begin{algorithm}[t]
\caption{Training Structured Recurrent SNN}
\label{alg:training}
\begin{algorithmic}[1]
\Require
Training data $\mathcal{D}$;
layers $\{L_k\}_{k=1}^K$ with sizes $\{N_k\}$;
bandwidth $b$; shortcuts $s$;
readout density $\rho_{\mathrm{out}}$;
learning rates $\eta_{\mathrm{out}}, \eta$;
decay factors $\lambda_e, \lambda_m$;
timesteps $T$
\Ensure
Trained synaptic weights

\State Initialize structured recurrent weights
\State Initialize sparse readout projections
\State Initialize fixed feedback matrices $B_k$
\State Initialize eligibility traces and modulatory states

\For{each epoch}
    \For{each mini-batch $(x,y)$}
        \State Encode $x$ into spikes $\{S^0(t)\}_{t=1}^T$
        
        \For{$t=1$ to $T$}
            \For{each layer $L_k$}
                \State Update neuron states and spikes
                \State Update eligibility traces
            \EndFor
        \EndFor
        
        \State Compute readout spike counts $z$
        \State Compute WTA error $\boldsymbol{\delta}$
        
        \State Update readout weights
        
        \For{each layer $L_k$}
            \State Compute broadcast feedback
            \State Update modulatory signal
            \State Apply synaptic plasticity
        \EndFor
        
        \State Update firing thresholds
    \EndFor
\EndFor
\end{algorithmic}
\end{algorithm}

\subsection{Computational Complexity}

Let $N = \sum_k N_k$ be the total number of neurons, $E$ the total number of synapses, $\bar{S}$ the average number of spikes per timestep, and $d$ the average out-degree. Event-driven inference has computational cost
\begin{equation}
\mathcal{C}_{\mathrm{infer}} = \mathcal{O}(T \bar{S} d),
\end{equation}
while training adds eligibility trace and plasticity updates of the same order:
\begin{equation}
\mathcal{C}_{\mathrm{train}} = \mathcal{O}(T \bar{S} d).
\end{equation}
Memory requirements scale as $\mathcal{O}(N)$ for neuron states, $\mathcal{O}(E)$ for synapses, and $\mathcal{O}(\rho E)$ for eligibility traces, where $\rho$ denotes the fraction of plastic synapses. Modulatory populations incur negligible additional memory overhead.

\subsection{Hardware Feasibility}

The proposed architecture is designed to align with the constraints of neuromorphic processors and FPGA-based accelerators by emphasizing sparse, event-driven computation, local state updates, and low-bandwidth global communication. Network execution is fully event-driven, such that computation and memory access occur only in response to spike events, allowing inference and learning cost to scale with spike activity rather than network size. This execution model is well suited for asynchronous neuromorphic systems such as Intel Loihi and SpiNNaker, where energy efficiency is achieved through sparse spike-based routing.

Connectivity is sparse by construction. Locally dense recurrent microcircuits are confined within layers, while long-range communication is realized through a small fraction of fixed small-world projections to the readout population. This structure limits fan-in and fan-out, reduces interconnect congestion, and simplifies address-event routing tables, which is critical for scalable deployment on both neuromorphic fabrics and FPGA interconnects.

Learning relies exclusively on local state variables. Eligibility traces are maintained per synapse using only pre- and postsynaptic spike information and can be implemented using local registers or accumulators colocated with synaptic memory. Feedback signals are low dimensional and low bandwidth: output errors are broadcast through fixed random projections and integrated by small modulatory populations that gate plasticity without injecting currents into neuronal dynamics. This avoids dense error backpropagation, weight symmetry, and global synchronization.

Because connectivity graphs remain fixed during training, only synaptic weights are updated online, further simplifying routing and memory management. Optional limited structural rewiring, if enabled, can be performed at slow timescales without disrupting real-time inference. Together, these properties make the proposed framework compatible with neuromorphic platforms such as Loihi and SpiNNaker, as well as time-multiplexed FPGA implementations, providing a practical pathway toward scalable hardware-compatible learning in recurrent spiking neural networks.

\section{Experimental Evaluation}

\subsection{Experimental Setup}

We evaluate the proposed architecture on the MNIST handwritten digit classification benchmark. Input images are encoded as Poisson spike trains with a fixed simulation horizon of $T=100$ timesteps. All experiments use a batch size of $128$ and are trained for $30$ epochs. Unless otherwise specified, the network consists of three structured recurrent layers with neuron counts $[4096,2048,4096]$, followed by a linear readout population with $C=10$ output units.

Within each recurrent layer, locally dense recurrence is enforced using a fixed bandwidth parameter $b=12$, while sparse long-range recurrent shortcuts are added to induce small-world connectivity. Each layer also maintains sparse long-range projections to the readout population. Synaptic weights are initialized randomly and trained using the proposed WTA teaching signal, broadcast alignment feedback, and modulatory plasticity with eligibility traces. No surrogate gradients, backpropagation, or backpropagation-through-time are used at any stage.

\subsection{Evaluation Metric}

Performance is reported using classification accuracy on the MNIST test set. Since learning is driven by a margin-based WTA teaching signal rather than likelihood maximization, cross-entropy loss is reported only for monitoring purposes and is not optimized during training. 

\begin{table*}[ht]
\centering
\caption{Comparison with non-gradient spiking neural network approaches on MNIST.}
\label{tab:baseline_comparison}
\begin{tabular}{l c c c}
\hline
Method & Learning rule & Depth / Recurrence & Test acc. (\%) \\
\hline
STDP \cite{Diehl2015} & Unsupervised STDP + readout & Shallow, feedforward & $\sim$95 \\
Reservoir SNN + readout & Fixed recurrence & Recurrent, untrained & 85--90 \\
Reward-modulated STDP & Three-factor plasticity & Shallow / recurrent & 88--90 \\
WTA-based SNN & Margin / competition & Shallow & 88--91 \\
e-prop \cite{Bellec2020} & Approx. gradient descent & Deep recurrent & 92--95$^\dagger$ \\
\hline
\textbf{This work} & \textbf{WTA + modulatory plasticity} & \textbf{Deep recurrent} & \textbf{$\sim$97} \\
\hline
\end{tabular}
\\[2pt]
{\footnotesize $^\dagger$e-prop approximates gradient descent and is included for reference only.}
\end{table*}

\subsection{Main Results}

Across all configurations, the proposed model exhibits stable learning dynamics and competitive performance despite sparse connectivity and purely local learning rules. For the best-performing configuration, the network achieves a peak test accuracy of approximately $\textbf{97.12}\%$, corresponding to a test error below $2.88\%$, using a readout density of $1.0$ for intermediate layers and $1.0$ for the final layer. Training remains stable over long runs, with no catastrophic divergence despite recurrent plasticity and sparse feedback.

These results demonstrate that deep structured recurrent SNNs can be trained effectively on MNIST using purely local, non-gradient learning mechanisms, achieving accuracy comparable to other non-surrogate-gradient SNN approaches reported in the literature.

To contextualize the proposed approach, Table~\ref{tab:baseline_comparison} compares our results with representative spiking neural network methods that do not rely on surrogate gradients or backpropagation. While gradient-based and ANN-to-SNN conversion methods can achieve higher absolute accuracy on MNIST, purely local learning rules without gradient optimization typically operate in the 85--95\% range. The proposed method achieves the state-of-the-art performance within this regime while supporting deep recurrent architectures, sparse connectivity, and low-bandwidth feedback.

\subsection{Effect of Readout Sparsity}

\begin{figure}
    \centering
    \includegraphics[width=0.8\linewidth]{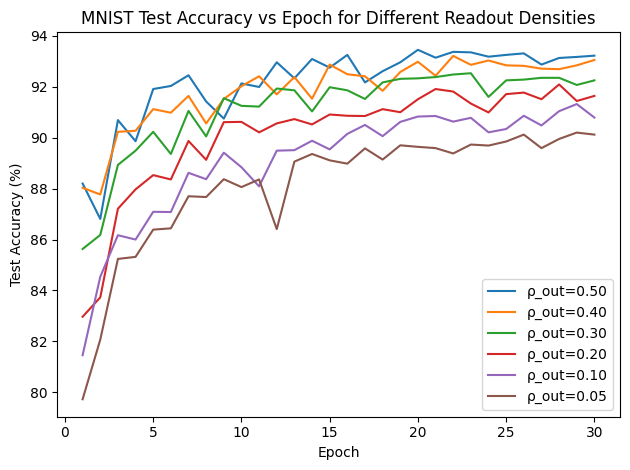}
    \caption{Test accuracy over training epochs for different readout density.}
    \label{fig:reachout_density}
\end{figure}

We first examine the impact of readout sparsity by varying the fraction of long-range projections from recurrent layers to the readout population, while keeping all other hyperparameters fixed. In our ablation study, we use a smaller neural networks with three layers of [1024 521 1024], and a final readout density of 0.5. Figure~\ref{fig:reachout_density} shows test accuracy as a function of training epoch for different readout densities, and Table~\ref{tab:readout_density} summarizes the final test accuracy obtained for different readout densities.

\begin{table}[h]
\centering
\caption{Effect of readout sparsity on MNIST test accuracy.}
\label{tab:readout_density}
\begin{tabular}{c c}
\hline
Readout density $\rho_{\mathrm{out}}$ & Test accuracy (\%) \\
\hline
0.50 & $\approx 93.4$ \\
0.40 & $\approx 93.0$ \\
0.30 & $\approx 92.3$ \\
0.20 & $\approx 91.6$ \\
0.10 & $\approx 91.3$ \\
0.05 & $\approx 90.1$ \\
\hline
\end{tabular}
\end{table}

As readout density decreases, performance degrades gradually rather than abruptly. Even with only $5\%$ of possible readout connections present, the network retains test accuracy above $90\%$. This highlights the robustness of the proposed learning framework to aggressive sparsification of long-range connections and supports its suitability for hardware-constrained implementations.

\subsection{Bandwidth and Shortcut Ablations}

Additional experiments varying recurrent bandwidth and the number of recurrent shortcuts, as shown in Table \ref{tab:ablation}, confirm the complementary roles of local recurrence and sparse long-range connectivity. Increasing bandwidth improves temporal mixing within layers but may reduce stability when excessively large, while adding a moderate number of shortcuts consistently improves convergence speed and final accuracy by reducing effective path length within recurrent layers. These observations are consistent with the intended small-world design of the recurrent connectivity.

\begin{table}[t]
\centering
\caption{Effect of bandwidth and shortcut.}
\label{tab:ablation}
\begin{tabular}{cccc}
\toprule
\multicolumn{2}{c}{Bandwidth} & \multicolumn{2}{c}{Shortcut} \\
\cmidrule(lr){1-2} \cmidrule(lr){3-4}
$b$ & Test accuracy (\%) & $s$ & Test accuracy (\%)\\
\midrule
6 & 92.3 & 128 & 92.8 \\
12 & 93.4 & 256 & 93.4 \\
24 & 94.02 & 512 & 94.15 \\
\bottomrule
\end{tabular}
\end{table}

\subsection{Discussion}

Overall, the experimental results demonstrate that structured local recurrence, sparse small-world communication, and modulatory learning signals together enable deep recurrent SNNs to learn supervised tasks using purely local plasticity rules. The graceful degradation observed under increasing sparsity and the absence of training instabilities underscore the robustness and scalability of the proposed approach, particularly in settings where gradient-based learning and dense feedback are impractical.

\section{Conclusion}

We demonstrate that deep structured recurrent SNNs can be trained using biologically grounded learning rules combining WTA teaching, broadcast alignment, and modulatory populations. The approach scales to thousands of neurons, preserves sparsity, and is compatible with neuromorphic hardware constraints. This framework provides a promising path toward scalable, energy-efficient learning systems beyond backpropagation.


\bibliographystyle{IEEEtran}
\bibliography{ref}

@article{Maass1997,
  title={Networks of spiking neurons: The third generation of neural network models},
  author={Maass, Wolfgang},
  journal={Neural Networks},
  volume={10},
  number={9},
  pages={1659--1671},
  year={1997}
}

@article{Indiveri2015,
  title={Memory and information processing in neuromorphic systems},
  author={Indiveri, Giacomo and Liu, Shih-Chii},
  journal={Proceedings of the IEEE},
  volume={103},
  number={8},
  pages={1379--1397},
  year={2015}
}

@article{Boahen2017,
  title={A neuromorph's prospectus},
  author={Boahen, Kwabena},
  journal={Computing in Science \& Engineering},
  volume={19},
  number={2},
  pages={14--28},
  year={2017}
}

@article{Neftci2019,
  title={Surrogate gradient learning in spiking neural networks},
  author={Neftci, Emre O and Mostafa, Hesham and Zenke, Friedemann},
  journal={IEEE Signal Processing Magazine},
  volume={36},
  number={6},
  pages={61--63},
  year={2019}
}

@article{Bellec2020,
  title={A solution to the learning dilemma for recurrent networks of spiking neurons},
  author={Bellec, Guillaume and Scherr, Felix and Subramoney, Anand and Hajek, Elias and Salaj, Darjan and Legenstein, Robert and Maass, Wolfgang},
  journal={Nature Communications},
  volume={11},
  number={1},
  pages={3625},
  year={2020}
}

@article{Diehl2015,
  title={Unsupervised learning of digit recognition using spike-timing-dependent plasticity},
  author={Diehl, Peter U and Cook, Matthew},
  journal={Frontiers in Computational Neuroscience},
  volume={9},
  pages={99},
  year={2015}
}

@article{Maass2002,
  title={Real-time computing without stable states: A new framework for neural computation based on perturbations},
  author={Maass, Wolfgang and Natschl{\"a}ger, Thomas and Markram, Henry},
  journal={Neural Computation},
  volume={14},
  number={11},
  pages={2531--2560},
  year={2002}
}

@article{Watts1998,
  title={Collective dynamics of small-world networks},
  author={Watts, Duncan J and Strogatz, Steven H},
  journal={Nature},
  volume={393},
  number={6684},
  pages={440--442},
  year={1998}
}

@article{Fremaux2016,
  title={Neuromodulated spike-timing-dependent plasticity, and theory of three-factor learning rules},
  author={Fr{\'e}maux, Nicolas and Gerstner, Wulfram},
  journal={Frontiers in Neural Circuits},
  volume={9},
  pages={85},
  year={2016}
}

@article{Gerstner2018,
  title={Eligibility traces and plasticity on behavioral time scales: Experimental support of neo-Hebbian three-factor learning rules},
  author={Gerstner, Wulfram and Lehmann, Matthias and Liakoni, Vasiliki and Corneil, Brian D and Brea, Johannes},
  journal={Frontiers in Neural Circuits},
  volume={12},
  pages={53},
  year={2018}
}

@article{Lillicrap2016,
  title={Random synaptic feedback weights support error backpropagation for deep learning},
  author={Lillicrap, Timothy P and Cownden, Daniel and Tweed, Douglas B and Akerman, Colin J},
  journal={Nature Communications},
  volume={7},
  pages={13276},
  year={2016}
}

@article{roelfsema2018control,
  title={Control of synaptic plasticity in deep cortical networks},
  author={Roelfsema, Pieter R and Holtmaat, Anthony},
  journal={Nature Reviews Neuroscience},
  volume={19},
  number={3},
  pages={166--180},
  year={2018},
  publisher={Nature Publishing Group UK London}
}

@book{GerstnerKistler2002,
  title={Spiking Neuron Models},
  author={Gerstner, Wulfram and Kistler, Werner M},
  publisher={Cambridge University Press},
  year={2002}
}

@article{Jaeger2001,
  title={The ``echo state'' approach to analysing and training recurrent neural networks},
  author={Jaeger, Herbert},
  journal={GMD Report 148},
  year={2001}
}

@article{BiPoo1998,
  title={Synaptic modifications in cultured hippocampal neurons: dependence on spike timing, synaptic strength, and postsynaptic cell type},
  author={Bi, Guo-qiang and Poo, Mu-ming},
  journal={Journal of Neuroscience},
  volume={18},
  number={24},
  pages={10464--10472},
  year={1998}
}

@article{Maass2000,
  title={On the computational power of winner-take-all},
  author={Maass, Wolfgang},
  journal={Neural Computation},
  volume={12},
  number={11},
  pages={2519--2535},
  year={2000}
}

\end{document}